# An ADMM-Incorporated Latent Factorization of Tensors Method for QoS Prediction

Jiajia Mi, Hao Wu

***Abstract***—As the Internet developed rapidly, it is important to choose suitable web services from a wide range of candidates. Quality of service (QoS) describes the performance of a web service dynamically with respect to the service requested by the service consumer. Moreover, the latent factorization of tenors (LFT) is very effective for discovering temporal patterns in high dimensional and sparse (HiDS) tensors. However, current LFT models suffer from a low convergence rate and rarely account for the effects of outliers. To address the above problems, this paper proposes an Alternating direction method of multipliers (ADMM)-based Outlier-Resilient Nonnegative Latent-factorization of Tensors model. We maintain the non-negativity of the model by constructing an augmented Lagrangian function with the ADMM optimization framework. In addition, the Cauchy function is taken as the metric function to reduce the impact on the model training. The empirical work on two dynamic QoS datasets shows that the proposed method has faster convergence and better performance on prediction accuracy.

## I. INTRODUCTION

AS THE INTERNET INFRASTRUCTURE IS UPGRADED ITERATIVELY [1][2], an increasing trend of enterprises are relocating their business to cloud service platforms. Given the different IT services supplied by various service providers, how should service consumers be able to select the most suitable IT service for their demand precisely? It is a hot issue that deserves our attention and solves [3]-[5][8].

Quality of service (QoS) is a non-functional characteristic, such as ports of call, throughput, and response time, which measures the performance of a Web service in answering requests [6]-[7]. In general, user-viewed QoS values have their personal characteristics, i.e., the sense of experience (QoS value) from different users consuming the same Web service has generally different. Therefore, it is important to obtain personalized QoS values from different users' choices, due to the challenges of: a) A user is unlikely to invoke all Web services offered by the providers that are able to get the complete QoS dataset. b) The QoS value getting from the customer who uses the same service offered by the equivalent provider would be changing over time. c) QoS datasets are collected from truth using experience, so capturing complete data in a long time period is time-consuming and costly [9]-[10]. Hence, it is essential to precisely predict missing QoS values based on observed values in order to achieve an estimate of personalized QoS values.

Researchers have proposed a wide range of solutions for the above issues, where latent factor analysis (LFA)-based ones [8]-[11] are increasingly catching on. Its strength is that LFA generates approximate low-rank matrices based on a few known values through various optimal techniques [12]-[19], i.e. LFA can efficiently deal with high-dimensional sparse matrix problems. Thus an LFA-based QoS predictor is able to correctly capture the correlations embedded in the known data and efficiently predict the values based on the observed QoS data through a well-designed learning algorithm. Although the LFA analysis model is efficient with high predictive accuracy, it does not take into account the dynamic changes in QoS data over time.

In order to tackle dynamic QoS prediction issues, an LFT-based QoS predictor is built which mostly uses the $L_2$-norm as loss function to measure the difference between prediction QoS value and observed one [13][20]-[21]. However, it is well-known that $L_2$-norm is exceptionally sensitive to outliers that is widespread in real application data [18][24]-[26]. Comparing with $L_2$-norm, the $L_1$-norm is more robust in dealing with the outliers, while it is hard for us to solve the optimization problem of the $L_1$-norm-based loss function [22]-[23]. As prior research [16][25], Cauchy loss performs robustly than $L_1$ or $L_2$ norm function when dealing with outliers. Therefore, this paper adopts the Cauchy loss to build learning objective and incorporates the optimization framework of the alternating directional multiplier method (ADMM) to achieve high prediction accuracy.

In overview, this paper has three main contributions as follows:

1) We propose a robust QoS prediction method that reduces the effect of outliers, where Cauchy loss is used as a distance function to improve model robustness.

2) We incorporate the ADMM method into LFT to provide an efficient solution for the QoS data prediction.

3) We perform experiments on two dynamic QoS datasets to evaluate the prediction accuracy and convergence speed of our method.

The rest of the paper is organized as follows: Section 2 introduces the preliminaries, Section 3 proposes our method, Section 4 depicts our experimental results in detail, and the last section concludes this paper.

---

✧ J. J. Mi and H. Wu are with the School of Computer Science and Technology, Dongguan University of Technology, Dongguan, Guangdong 523808, China (e-mail: Mja_ii@163.com, haowuf@gmail.com).

## II. PRELIMINARIES

### A. QoS Data Tensor

We use the user-service-time tensor as the input data source, as shown in Fig. 1. Due to the unique features of the QoS dataset and the real-world significance that contains only a few non-negative known elements (i.e. $|\Lambda| << |\Gamma|$), the target tensor is usually non-negative high-dimensional sparse and is defined as follows:

***Definition* 1:** (*HiDS User-Service-Time Tensor*): Given a tensor $\mathbf{Y}^{|I|\times|J|\times|K|}$ denotes a user-service-time tensor, in which element $y_{ijk}$ represents the QoS scored by user $i \in I$ on service $j \in J$ at the time-point $k \in K$.

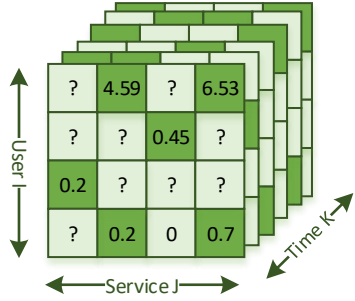

Fig. 1. An example of user-service-time matrix for response time.

### B. Problem Formulation

In this paper, we apply Canonical/Polyadic Tensor Factorization (CPTF) to performs the latent factorization of tensors on $\mathbf{Y}$ [27][28]. In terms of CPTF, we decompose tensor $\mathbf{Y}$ into $R$ rank-one tensors like $\mathbf{A}_1, \mathbf{A}_2, \ldots, \mathbf{A}_R$, in which $R$ is the rank of approximation tensor $\hat{\mathbf{Y}}$.

***Definition* 2:** (Rank-One Tensor): $\mathbf{A}_r^{|I|\times|J|\times|K|}$ is a rank-one tensor which could be calculated by three latent feature (LF) vectors $\boldsymbol{u}_r, \boldsymbol{s}_r,$ and $\boldsymbol{t}_r$ as $\mathbf{A}_r = \boldsymbol{u}_r \circ \boldsymbol{s}_r \circ \boldsymbol{t}_r$.

The LF matrices U, S, and T are made of $R$ LF vectors with $|I|$, $|J|$ and $|K|$ particularly, in which are combined to form the $r$ th rank-one vector $\mathbf{A}_r$ [29]-[30]. We could achieve the element-wise expression of $\mathbf{A}_r$ by expanding the outer product of the three vectors.

$$a_{ijk} = u_{ir} s_{jr} t_{kr}. \tag{1}$$

As for these $R$ rank-one tensors, i.e. $\{\mathbf{A}_r \mid r \in \{1, 2, \ldots, R\}\}$, we obtain the rank-$R$ approximation $\hat{\mathbf{Y}}$ of $\mathbf{Y}$ as follows:

$$\hat{y}_{ijk} = \sum_{r=1}^{R} \mathbf{A}_{ijk}^{(r)} = \sum_{r=1}^{R} u_{ir} s_{jr} t_{kr}. \tag{2}$$

For achieving the required LF matrices U, S, and T, it is common to use Euclidean distance as a similarity measure function measure the distance between $\hat{\mathbf{Y}}$ and $\mathbf{Y}$ in LFT. Note that just a few entries of $\mathbf{Y}$ is known for us, the objective function should define on known data set $\Lambda$ for a precise prediction, and it is necessary to introduce non-negative condition for U, S, and T for their real meaning. Thus, the explicit objective function can be clearly given as follows:

$$\varepsilon = \sum_{y_{ijk} \in \Lambda} \left(y_{ijk} - \hat{y}_{ijk}\right)^2 = \sum_{y_{ijk} \in \Lambda} \left(y_{ijk} - \sum_{r=1}^{R} u_{ir} s_{jr} t_{kr}\right)^2, \tag{3}$$

$$s.t. \ \forall i \in I, \forall j \in J, \forall k \in K, r \in \{1, \ldots, R\}:$$

$$u_{ir} \geq 0, s_{jr} \geq 0, \ t_{kr} \geq 0.$$

To deal with scale differences between instances, introducing linear biases for objection function [9]. Based on [31]-[33], the linear bias of the target third-order tensor $\mathbf{Y}$ can be denoted by three linear bias vectors $\boldsymbol{a}, \boldsymbol{b},$ and $\boldsymbol{c}$ of length $|I|, |J|, |K|$. The further expansion of (3) is:

$$\varepsilon = \sum_{y_{ijk} \in \Lambda} \left(\left(y_{ijk} - \sum_{r=1}^{R} u_{ir} s_{jr} t_{kr} - a_i - b_j - c_k\right)^2\right),$$

$$s.t. \ \forall i \in I, j \in J, k \in K, r \in \{1, 2, \ldots, R\}: \tag{4}$$

$$u_{ir} \geq 0, s_{jr} \geq 0, t_{kr} \geq 0, \ a_i \geq 0, b_j \geq 0, c_k \geq 0.$$

## III. 4 OUR METHOD

### A. Objective Function

As mentioned before, both $L_1$-norm and $L_2$-norm are sensitive to outliers. It has been shown that the Cauchy loss is resistant to outliers [32]-[34]. To enable our method to be more robust to outliers, we take advantage of the Cauchy loss as a measure of the difference between the observed value of QoS and predicted ones, as:

$$\varepsilon = \sum_{y_{ijk} \in \Lambda} \left( \ln \left( 1 + \left( y_{ijk} - \sum_{r=1}^{R} u_{ir} s_{jr} t_{kr} - a_i - b_j - c_k \right)^2 \middle/ \gamma^2 \right) \right),$$
$$s.t. \ \forall i \in I, j \in J, k \in K, r \in \{1, 2, ..., R\}:$$
$$u_{ir} \geq 0, s_{jr} \geq 0, t_{kr} \geq 0, \ a_i \geq 0, b_j \geq 0, c_k \geq 0.$$
(5)

The ADMM-based models are typically coupled with augmented Lagrangian methods [35]. However, the non-negative constraints in (5) would lead to solving the corresponding augmented Lagrangian question into a hard-to-solve problem [36]. To address this problem, we inserted the auxiliary variables $\tilde{U}$, $\tilde{S}$, $\tilde{T}$, $\tilde{a}$, $\tilde{b}$, and $\tilde{c}$ respectively corresponding to U, S, T, *a*, *b*, and *c*, into the learning objective, thus adding new equation constraints. Introducing the augmented Lagrangian into this issue for ensuring the non-negativity of the latent features U, S, T, *a*, *b*, and *c*, as well as the Unconstrained optimization of $\varepsilon$:

$$L(\Phi) = \frac{1}{2} \sum_{y_{ijk} \in \Lambda} \left( \ln \left( 1 + \left( y_{ijk} - \sum_{r=1}^{R} \tilde{u}_{ir} \tilde{s}_{jr} \tilde{t}_{kr} - \tilde{a}_i - \tilde{b}_j - \tilde{c}_k \right)^2 \middle/ \gamma^2 \right) \right)$$
$$+ \sum_{(i,r)} \frac{\tau_i}{2} (\tilde{u}_{ir} - u_{ir} + \phi_{ir}/\tau_i)^2 + \sum_{(j,r)} \frac{\upsilon_j}{2} (\tilde{s}_{jr} - s_{jr} + \rho_{jr}/\upsilon_j)^2 + \sum_{(k,r)} \frac{\omega_k}{2} (\tilde{t}_{kr} - t_{kr} + \psi_{kr}/\omega_k)^2$$
$$+ \sum_{i} \frac{\alpha_i}{2} (\tilde{a}_i - a_i + \chi_i/\alpha_i)^2 + \sum_{j} \frac{\beta_j}{2} (\tilde{b}_j - b_j + \varphi_j/\beta_j)^2 + \sum_{k} \frac{\delta_k}{2} (\tilde{c}_k - c_k + \sigma_k/\delta_k)^2 - \vartheta.$$
(6)

Where $\vartheta$ is given as:
$$\vartheta = \sum_{(i,r)} \phi_{ir}^2 / 2\tau_i + \sum_{(j,r)} \rho_{jr}^2 / \upsilon_j + \sum_{(k,r)} \psi_{kr}^2 / \omega_k + \sum_{i} \chi_i^2 / \alpha_i + \sum_{j} \varphi_j^2 / \beta_j + \sum_{k} \sigma_k^2 / \delta_k$$
(7)

Furthermore, since $\Lambda$ has data imbalances, i.e. user *i* (or service *j*, time slice *k*) is associated with known elements in **Y**, having different distributions, the augmentation constants require careful selection to precisely control augmentation effect from the perspective of data density. Specifically, they are given as[38]:

$$\tau_i = \alpha_i = \lambda |\Lambda(i)|, \upsilon_j = \beta_j = \lambda |\Lambda(j)|, \omega_k = \delta_k = \lambda |\Lambda(k)|.$$
(8)

*B. Learning Scheme*

*1) Auxiliary Variable:* We take into account the alternating learning strategy [35] for (6), namely fixing the other variables in (6) and solving the optimization problem for the auxiliary variable. Thus (6) becomes a convex problem with an analytic solution. i.e. The optimal values of the auxiliary variable $\tilde{u}_{ir}$ is derived as follows:

$$\frac{\partial J}{\partial \tilde{u}_{ir}} = -\sum_{y_{ijk} \in \Lambda(i)} \Delta_{ijk} \cdot \tilde{s}_{jr} \tilde{t}_{kr} (y_{ijk} - \hat{y}_{ijk}) + \tau_i (\tilde{u}_{ir} - u_{ir} + \phi_{ir}/\tau_i)$$
$$= -\sum_{y_{ijk} \in \Lambda(i)} \Delta_{ijk} \cdot \tilde{s}_{jr} \tilde{t}_{kr} (y_{ijk} - \hat{y}_{ijk}^r) + \sum_{y_{ijk} \in \Lambda(i)} \Delta_{ijk} \cdot \tilde{u}_{ir} (\tilde{s}_{jr} \tilde{t}_{kr})^2 + \tau_i \tilde{u}_{ir} - \tau_i u_{ir} + \phi_{ir}.$$
(9)

where $\hat{y}_{ijk}^r = \sum_{n=1, n \neq r}^{R} \tilde{u}_{in} \tilde{s}_{jn} \tilde{t}_{kn} + \tilde{a}_i + \tilde{b}_j + \tilde{c}_k$ and $\Delta_{ijk}$ is defined as $\Delta_{ijk} = 1 / (\gamma^2 + (y_{ijk} - \hat{y}_{ijk})^2)$. Making $(\partial L / \partial \tilde{u}_{ir}) = 0$, we get:

$$\tilde{u}_{ir} = \frac{\sum_{y_{ijk} \in \Lambda(i)} \Delta_{ijk} \cdot \tilde{s}_{jr} \tilde{t}_{kr} (y_{ijk} - \hat{y}_{ijk}^r) + \tau_i u_{ir} - \phi_{ir}}{\tau_i + \sum_{y_{ijk} \in \Lambda(i)} \Delta_{ijk} \cdot (\tilde{s}_{jr} \tilde{t}_{kr})^2}$$
(10)

As well, we can obtain the update rule of $\tilde{s}_{jr}$ and $\tilde{t}_{kr}$ as:

$$\tilde{s}_{jr} = \frac{\sum_{y_{ijk} \in \Lambda(j)} \Delta_{ijk} \cdot \tilde{u}_{ir} \tilde{t}_{kr} (y_{ijk} - \hat{y}_{ijk}^r) + \upsilon_j s_{jr} - \rho_{jr}}{\upsilon_j + \sum_{y_{ijk} \in \Lambda(j)} \Delta_{ijk} \cdot (\tilde{u}_{ir} \tilde{t}_{kr})^2},$$

$$\tilde{t}_{kr} = \frac{\sum_{y_{ijk} \in \Lambda(k)} \Delta_{ijk} \cdot \tilde{u}_{ir} \tilde{s}_{jr} (y_{ijk} - \hat{y}_{ijk}^r) + \omega_k t_{kr} - \psi_{kr}}{\omega_k + \sum_{y_{ijk} \in \Lambda(k)} \Delta_{ijk} \cdot (\tilde{u}_{ir} \tilde{s}_{jr})^2}.$$
(11)

Equivalently, we also can achieve the update rule of {$\tilde{a}$, $\tilde{b}$, $\tilde{c}$} as:

$$\tilde{a}_i = \frac{\sum_{y_{ijk} \in \Lambda(i)} \Delta_{ijk} \cdot \left( y_{ijk} - \left( \sum_{r=1}^{R} \tilde{u}_{ir}\tilde{s}_{jr}\tilde{t}_{kr} + \tilde{b}_j + \tilde{c}_k \right) \right) + \alpha_i a_i - \chi_i}{\alpha_i + \Delta_{ijk} \cdot |\Lambda(i)|},$$

$$\tilde{b}_j = \frac{\sum_{y_{ijk} \in \Lambda(j)} \Delta_{ijk} \cdot \left( y_{ijk} - \left( \sum_{r=1}^{R} \tilde{u}_{ir}\tilde{s}_{jr}\tilde{t}_{kr} + \tilde{a}_i + \tilde{c}_k \right) \right) + \beta_j b_j - \varphi_j}{\beta_j + \Delta_{ijk} \cdot |\Lambda(j)|}, \quad (12)$$

$$\tilde{c}_k = \frac{\sum_{y_{ijk} \in \Lambda(k)} \Delta_{ijk} \cdot \left( y_{ijk} - \left( \sum_{r=1}^{R} \tilde{u}_{ir}\tilde{s}_{jr}\tilde{t}_{kr} + \tilde{a}_i + \tilde{b}_j \right) \right) + \delta_k c_k - \sigma_k}{\delta_k + \Delta_{ijk} \cdot |\Lambda(k)|}.$$

*2) Latent Features:* Analogously, with the potential features {U, S, T} and {*a*, *b*, *c*} updated, the output latent features are projected to the nonnegativity real field to satisfy the non-negative constraints as:

$$u_{ir} = \max(0, \tilde{u}_{ir} + \tau_i/\phi_{ir}), s_{jr} = \max(0, \tilde{s}_{jr} + \upsilon_j/\rho_{jr}), t_{kr} = \max(0, \tilde{t}_{kr} + \omega_k/\psi_{kr});$$
$$a_i = \max(0, \tilde{a}_i + \alpha_i/\chi_i), b_j = \max(0, \tilde{b}_j + \beta_j/\varphi_j), c_k = \max(0, \tilde{c}_k + \delta_k/\sigma_k). \quad (13)$$

*3) Lagrangian Multipliers:* Subsequently, taken dual gradient ascent on the Lagrangian multiplies as [37]:

$$\phi_{ir} = \phi_{ir} + \eta\tau_i(\tilde{u}_{ir} - u_{ir}), \rho_{jr} = \rho_{jr} + \eta\upsilon_j(\tilde{s}_{jr} - s_{jr}), \psi_{kr} = \psi_{kr} + \eta\omega_k(\tilde{t}_{kr} - t_{kr});$$
$$\chi_i = \chi_i + \eta\alpha_i(\tilde{a}_i - a_i), \varphi_j = \varphi_j + \eta\beta_j(\tilde{b}_j - b_j), \sigma_k = \sigma_k + \eta\delta_k(\tilde{c}_k - c_k). \quad (14)$$

## IV. EXPERIMENTAL RESULTS AND DISCUSSION

### A. General Settings

*1) Evaluation Metrics:* In this study, we adopt Mean Absolute Error (MAE) as evaluation metrics for whether the model accurately captures the features underlying an HiDS tensor. If $\hat{y}_{ijk}$ and $y_{ijk}$ are the estimated and actual values respectively, then the two expressions can be written as follows:

$$\text{MAE} = \frac{1}{|\Lambda|} \sum_{y_{ijk} \in \Lambda} |y_{ijk} - y_{ijk}|. \quad (15)$$

*2) Datasets:* We conduct experiments in WS Monitor [39] on two publicly available datasets, which are dynamic datasets collected from real web services. It profiles 4500 web services' response times and throughput to 142 real user requests over 64 different time periods. Thus, we can yield two *User × Service × Time* tensors of size 142 × 4532 × 64, with 30,287,611 known QoS records. There were three different test ratios built separately for this experiment, as shown in Table I. In Table I, |M|: |N|: |O| denote the proportions of the training set, validation set and test set. For example, the |M|: |N|: |O| ratio for D1.1 is 16%:4%:80%, which indicates that we randomly selected 16% from D1 as the training set and 4% as the validation set to construct the model and tune the hyper-parameters. The remaining 80% of the D1.1 is predicted to evaluate its performance.

TABLE I. Detailed Settings of Testing Cases.

| Dataset | Case | \|M\|:\|N\|:\|O\| | \|M\| | \|N\| | \|O\| |
|---|---|---|---|---|---|
| D1 | D1.1 | 16%:4%:80% | 4,846,017 | 1,211,504 | 24,230,088 |
|  | D1.2 | 20%:5%:75% | 6,057,522 | 1,514,380 | 23,851,493 |
| D2 | D2.1 | 16%:4%:80% | 4,846,017 | 1,211,504 | 24,230,088 |
|  | D2.2 | 20%:5%:75% | 6,057,522 | 1,514,380 | 23,851,493 |

### B. Compared Models

In the next, we compare our model with the following four advanced models.

M1: a high dimension oriented QoS prediction model (HDOP) proposed in [39]. It exploits the structural relationships of multidimensional QoS data and model bases on the concept of multi linear algebra.

M2: a QoS predictor for service-oriented environments [40], which construct a multi-dimensional relationship between QoS evaluation factors and service provider QoS values based on the collected reference reports

M3: an NLFT model with learning-depth adjustable model [41], which investigate the link between a NLFT's performance and its learning depth for handling train fluctuation problem in most models.

M4: a state-of-the-art robust QoS prediction method proposed in [42]. It takes into account the effect of outliers and the experimental results show it has highly robust to outliers.

M5: an NLFT model based on Cauchy loss with ADMM optimization frame-work proposed in this paper.

### C. Experimental Results

The experimental result of M1-5 are depicted in Table II respectively. From these results, we can find that M5 gets higher prediction accuracy compared with its peers. As shown in Table II, M5 outperforms the other models in all testing cases for predicting the missing QoS data. For example, on D1.2, the MAE of M5 is 1.1107 with an accuracy improvement of 19.98%, 13.73%, 7.75% and 5.38% over M1's 1.3881, M2's 1.2875 M3's 1.2041, and M4's 1.1739, respectively. On D2.2, the MAE of M5 is 1.1107 with an accuracy improvement of 20.53%, 12%, 2.98%, and 6.41% over M1's 4.1961, M2's 3.7888 M3's 3.4366 and M4's 3.5627, respectively. We also can draw a similar conclusion on other cases shown in Table II.

TABLE II. Lowest MAE of Each Model on All Testing Case.

| Testing Case | M1 | M2 | M3 | M4 | M5 |
|---|---|---|---|---|---|
| D1.1 | 1.3978 | 1.3269 | 1.2263 | 1.1745 | 1.1650 |
| D1.2 | 1.3881 | 1.2875 | 1.2041 | 1.1739 | 1.1107 |
| D2.1 | 5.4885 | 4.0108 | 3.5131 | 3.5957 | 3.4193 |
| D2.2 | 4.1961 | 3.7888 | 3.4366 | 3.5627 | 3.3343 |

## V. CONCLUSION

In this paper, we propose an ADMM-based Outlier-Resilient Nonnegative Latent factorization of Tensors model. We focus on two aspects of model construction: First, we construct an augmented Lagrangian function to deal with non-negativity constrains, second, we use the Cauchy loss to metric the different between the true QoS value and the predicted QoS value to reduce the impact of outliers on model training. We conduct comparative experiments on two dynamic QoS datasets and the findings show that our approach has higher prediction accuracy.